\documentclass[conference]{IEEEtran}
\IEEEoverridecommandlockouts

\usepackage{cite}
\usepackage{amsmath,amssymb,amsfonts}
\usepackage{algorithmic}
\usepackage{graphicx}
\usepackage{textcomp}
\usepackage{xcolor}
\usepackage{multirow}
\usepackage{booktabs}
\usepackage{url}
\usepackage{tikz}
\usetikzlibrary{positioning,patterns,arrows.meta}
\def\BibTeX{{\rm B\kern-.05em{\sc i\kern-.025em b}\kern-.08em
    T\kern-.1667em\lower.7ex\hbox{E}\kern-.125emX}}

\def\etal{\emph{et al.}}

\newcommand{\VEC}[1]{\mathbf{#1}}
\newcommand{\VECG}[1]{\boldsymbol{#1}}
\newcommand{\R}{\mathbb{R}}
\newcommand{\N}{\mathbb{N}}
\newcommand{\I}{\mathbb{I}}

\newcommand{\sv}[1]{\textcolor{black}{#1}}

\newif\ifreview
\reviewfalse

\begin{document}

\title{Combining Foundation Model Confidence and Monocular Depth for Training-Free Out-of-Distribution Segmentation}

\ifreview
\author{\IEEEauthorblockN{Anonymous Authors}
\IEEEauthorblockA{Anonymous Institution(s)\\Paper ID: 493}
}
\else
\author{\IEEEauthorblockN{Serin Varghese}
\IEEEauthorblockA{\textit{Heinrich-Heine-University} \\
\textit{CARIAD SE}\\
Düsseldorf, Germany \\
serin.varghese@hhu.de}
\and
\IEEEauthorblockN{Fabian Hueger}
\IEEEauthorblockA{\textit{CARIAD SE} \\
Wolfsburg, Germany \\
fabian.hueger@cariad.technology}
\and
\IEEEauthorblockN{Kira Maag}
\IEEEauthorblockA{\textit{Heinrich-Heine-University} \\
Department of Computer Science\\
Düsseldorf, Germany \\
kira.maag@hhu.de}
}
\fi

\maketitle

\begin{abstract}
Autonomous vehicles operating in open-world scenarios are inevitably confronted with previously unknown objects, such as exotic animals or loose cargo. The reliable detection and segmentation of these out-of-distribution (OOD) objects is therefore crucial for a safe understanding of the environment and decision-making.
Most existing approaches require access to OOD training samples, retraining of the segmentation backbone, or dedicated auxiliary architectures, limiting their practical applicability. We propose a training-free method that derives dense OOD scores directly from the confidence predictions of a foundation segmentation model, without any task-specific fine-tuning or access to anomalous data.
To improve the robustness of our OOD segmentation, geometric information from monocular depth estimation is incorporated into the decision process, providing complementary cues to uncertainty-based predictions.
We evaluate the proposed method on the SegmentMeIfYouCan benchmark and additionally assess its performance on OOD tracking in video sequences, reflecting the temporal nature of real-world perception systems. \sv{The method performs strongly on road-centered benchmarks. 
}
\end{abstract}


\section{Introduction}
 
Deep neural networks (DNNs) have achieved remarkable performance in computer vision tasks such as object detection \cite{Zong2023} and semantic segmentation \cite{Cheng2022}.
While object detection localizes and classifies individual objects using bounding boxes, semantic segmentation provides pixel-wise scene understanding by assigning a semantic category to each image pixel. 
In safety-critical applications such as automated driving, semantic segmentation plays a central role in understanding the surrounding environment. 
Although modern segmentation models achieve high performance on established benchmarks \cite{Hummer2024}, they are based on the assumption that the observed objects belong to the predefined set of semantic classes that was available during training.
In open-world scenarios, however, automated vehicles may encounter previously unknown objects such as loose cargo, exotic animals, fallen vegetation, or other unexpected obstacles. 
Such out-of-distribution (OOD) objects can lead to overconfident and potentially dangerous predictions, especially when they appear on the road in front of the ego-vehicle \cite{Pinggera2016}. 
Consequently, the reliable segmentation of OOD objects has become a crucial research topic for perception systems \cite{Blum2019_1,Chan2021_1}. 
Since sensor data in automated driving is captured as video streams, identifying OOD objects in individual frames is often only the first step. 
To support downstream tasks such as risk assessment and motion planning, unknown objects must also be consistently localized over a longer time period. As a result, modern benchmarks take into account both the segmentation and tracking of unknown objects in image sequences \cite{Maag2022_OOD_tracking}.
\begin{figure}[t!]
    \centering
    \includegraphics[width=0.99\linewidth]{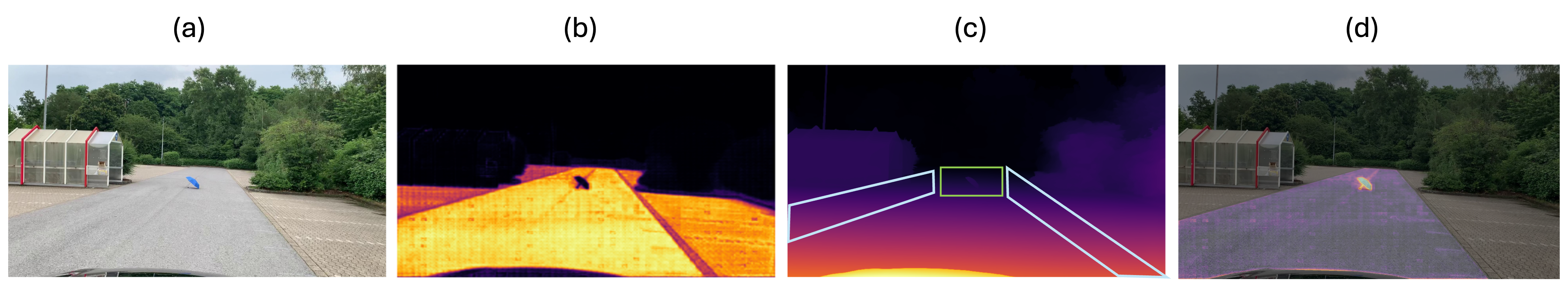}
    \caption{(a) Input image from the SOS dataset, (b) road confidence by the foundation segmentation model, (c) monocular depth estimation (the blue boxes show the smooth areas that are not OODs and can be suppressed), and (d) our obtained OOD score.}
    \label{fig:motivation}
\end{figure}

A wide range of methods for OOD segmentation has been proposed over the last years. 
Existing methods range from techniques for estimating uncertainty based on network outputs \cite{Hendrycks2016,Mukhoti2018}, through feature-space modeling \cite{Sodano2024,Vojir2024} and reconstruction-based approaches \cite{Vojir2021,Vojir2023}, to methods that incorporate OOD examples during training (whether real \cite{Holle2025,Zhong2024} or synthetic generated by generative models \cite{Delic2024,Won2026}). 
Among these approaches, uncertainty estimation is particularly interesting since it can often be directly applied to pretrained segmentation models and provides an intuitive measure of prediction reliability. Thus, these models do not require retraining, additional OOD datasets, or auxiliary model components. 
Most uncertainty-based OOD segmentation methods are based on traditional semantic segmentation networks. 
While these models achieve excellent results on their target task, they are typically trained on a fixed set of semantic categories and therefore suffer from the limitations of closed-set learning. 
In contrast, recent foundation models \cite{Ravi2025} are trained on significantly larger and more diverse datasets, enabling them to capture a broader spectrum of visual concepts and generalize more robustly to previously unseen objects and environments. 
This allows them to provide complementary information that is not captured by the uncertainty estimates of traditional segmentation networks alone.

\sv{In this work, we introduce a training-free approach for road-centered OOD segmentation (and a preliminary tracking extension) that combines uncertainty information from a foundation model with monocular depth estimates.}
Our key finding is that modern foundation segmentation models, such as SAM\,2~\cite{Ravi2025}, already provide valuable information for OOD detection, even without explicit OOD supervision. 
Instead of directly predicting anomalies, we use the model to robustly segment the drivable road area and interpret regions with low prediction confidence within the road mask as potential OOD objects. 
\sv{We use road segmentation as a proxy because OOD objects have neither a known identity nor a universal location for a reliable anomaly prompt. The road is a stable target whose deviations reveal unexpected objects without prior appearance knowledge.}
Thus, the model does not require OOD-specific training data or additional retraining, but rather leverages the strong generalization capabilities acquired through large-scale pretraining on diverse data. 
Furthermore, the segmentation process can be flexibly controlled using various prompt types, including positive and negative point prompts as well as bounding boxes. However, a simple point prompt is insufficient, as we demonstrate experimentally. 
Despite the high robustness of the segmentation model, the resulting confidence predictions still tend to generate false positives in visually complex areas such as textured road surfaces or semantic scene boundaries.
To address these false positives, we incorporate (i) classical image processing techniques to suppress small noise fragments and unstable boundary predictions, and (ii) monocular depth estimation \cite{Yang2024} as an additional geometric consistency cue. 
The underlying assumption is that true OOD obstacles exhibit a measurable depth difference relative to the surrounding road surface, whereas many false positives, such as textured asphalt, lane markings, or manhole covers, remain approximately coplanar. 
We therefore perform a segment-level verification of connected OOD components by comparing their estimated depth to the surrounding road context. 
Components with sufficiently distinct depth are retained as valid OOD detections, while geometrically consistent regions are suppressed as false positives. 
Furthermore, since autonomous driving systems operate on continuous video frames, we extend the proposed framework to OOD tracking by associating validated OOD objects across consecutive frames.
\ifreview The source code is omitted for double-blind review. \else \sv{The source code of our method is publicly available online.\footnote{\url{https://github.com/serin-varghese/OODTrack}}} \fi

Our contributions can be summarized as follows:
\begin{itemize}
    \item We propose a novel training-free OOD detection pipeline that derives dense anomaly scores from inverted SAM\,2 road segmentation logits, requiring only a fixed set of geometric point prompts and no task-specific fine-tuning.
    \item We introduce depth-ring scoring, a component-level fusion mechanism that exploits local depth disparity from monocular depth estimation to suppress false positives while preserving detection recall by construction.
    \item \sv{We demonstrate that our method achieves strong performance on road-centered OOD benchmarks, without task-specific training data, OOD examples, or task-specific auxiliary models. The resulting tracking component is a simple baseline for temporal association. 
    }
\end{itemize}

%
%
%
\section{Related Work}\label{sec:rel_work}
\subsection{OOD Segmentation}
For OOD segmentation, various methods for uncertainty estimation were investigated. 
Common approaches include confidence-based methods such as maximum softmax probability \cite{Hendrycks2016}, as well as sampling-based strategies such as MC Dropout \cite{Mukhoti2018} and deep ensembles \cite{Lakshminarayanan2017}. 
Subsequently, pixel-wise gradient norms (PGN) were introduced to capture the propagation of uncertainty during the network’s forward pass \cite{Maag2024_grads}, as well as Wasserstein-based evidential uncertainty (W-EDL) was applied \cite{Brosch2025}.
Other methods, such as Mahalanobis distance \cite{Lee2018} and ODIN \cite{Liang2018}, improve the separation between in-distribution and OOD samples by applying adversarial perturbations to the input images.

Beyond output-based uncertainty estimation, several approaches focus on the feature space of DNNs. 
One method estimates the density of feature representations within the distribution using nearest-neighbor techniques \cite{Galesso2023}, while another relies on online data condensation and low-dimensional projection spaces to derive calibrated OOD decision boundaries \cite{Vojir2024}. 
The method described in \cite{Sodano2024} applies multiple decoders to cluster features of the same semantic class while simultaneously performing anomaly segmentation. 
In \cite{Ackermann2023}, the raw mask predictions from mask-based semantic segmentation networks are used, whereby anomaly-related masks are typically discarded during standard semantic prediction, and in \cite{Marschall2025}, a multi-scale approach based on foreground-background segmentation models is presented.

Another research direction involves incorporating OOD data directly into the training process. This additional data are intentionally different from the original training distribution and can consist of either real images \cite{Biase2021,Chan2021,Gao2023,Grcic2022,Grcic2023,Holle2025,Liu2023,Rai2023,Nayal2024,Tian2022,Zhong2024} or synthetically generated examples. 
Synthetic negative examples, for instance, have been used to suppress the energy of anomalous pixels \cite{Nayal2023}. 
Another approach combines in-distribution uncertainty with explicit modeling of the negative class to derive improved outlier scores \cite{Delic2024}. 
Normalizing flows have also been used to synthesize negative examples for OOD-aware training \cite{Blum2019_1,Grcic2021,Gudovskiy2023,Won2026}. 


\begin{figure*}[t!]
	\centering
    \includegraphics[width=0.8\linewidth]{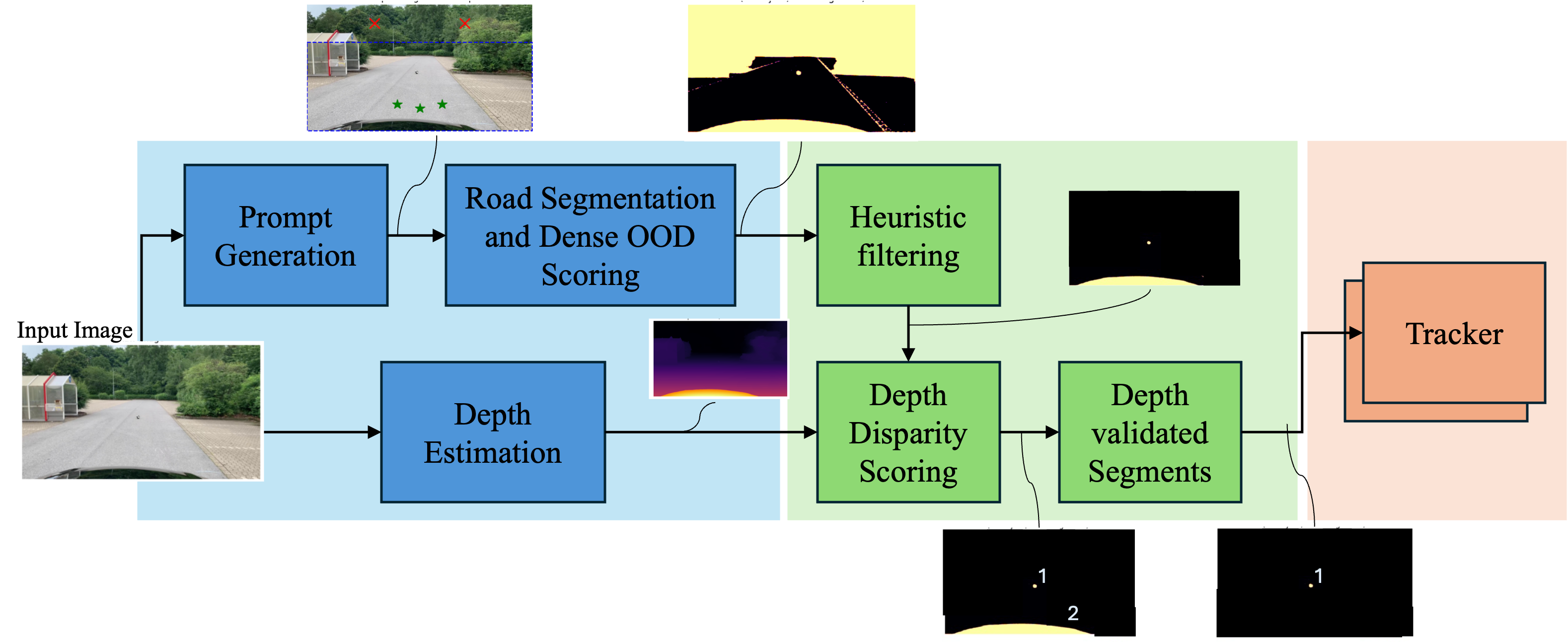}
	\caption{\textbf{Block diagram of the proposed depth-ring OOD detection pipeline.} The input image~$\VEC{x}_t$ is processed by two branches: SAM\,2 provides dense anomaly scores $\VEC{s}^{\mathrm{SAM}} \in \I^{H \times W}$ via inverted road segmentation logits and the road mask~$\mathcal{M}_{\mathrm{road}}$; Depth Anything~V2 estimates monocular depth~$\hat{\VEC{d}}$. Connected components~$\mathcal{C}_n$ from thresholded OOD scores are scored by \textbf{depth disparity}~$\Delta_n$ against a surrounding road ring~$\mathcal{R}_n$, retaining only components satisfying $\Delta_n \geq \delta_{\min}$.}
	\label{fig:overview}
\end{figure*}

In addition, several methods rely on auxiliary architectures or reconstruction-based frameworks. 
In \cite{Besnier2021}, input images are distorted through local adversarial attacks and an observer network is trained to predict segmentation errors. 
Other approaches compare original and resynthesized images using discrepancy networks to highlight unexpected objects \cite{Lis2019,Lis2020}. 
Reconstruction-based techniques \cite{Vojir2021,Vojir2023} model the normal appearance of street scenes and assume that flawed reconstructions indicate areas outside the training distribution. 
In \cite{Zhang2024}, multiple specialized components, such as a pixel decoder, a transformer decoder, a teacher network, and multilayer perceptrons, are combined into unified frameworks for OOD segmentation. 

\sv{In contrast to many existing OOD segmentation approaches, our method does not rely on additional OOD training data, the generation of synthetic negative examples, retraining procedures, or a task-specific auxiliary architecture.}
Instead, we build upon a segmentation model and leverage uncertainty information derived directly from its predictions. These uncertainty estimates are then combined with a depth prediction to improve OOD detection. Consequently, our work is conceptually closest to uncertainty-based OOD segmentation methods. 

\subsection{OOD Tracking}
Compared to OOD segmentation, OOD tracking in video sequences has received significantly less attention. Existing approaches typically combine an OOD segmentation method with a subsequent tracking phase. 
The Maximized Entropy method relies on additional OOD training data and links segmented objects together based on mask overlaps \cite{Chan2021}.
JSRNet \cite{Vojir2021} and DaCUP \cite{Vojir2023} use generative auxiliary models to estimate OOD regions, while PixOOD \cite{Vojir2024} perform OOD detection in the latent feature space. For temporal alignment, these three methods formulate tracking as a detection-matching problem and solve a linear matching optimization problem between object instances across consecutive frames. 
UNO-SAM2 \cite{Delic2024} uses synthetic negative examples for OOD segmentation and employs SAM2 for mask propagation and tracking. However, the tracking method uses information from both past and future frames, making it unsuitable for online deployment. 
Furthermore, a retrieval-based approach (RbA, \cite{Shoeb2024}) has been proposed that combines OOD segmentation with multi-modal foundation models to retrieve and group similar obstacle instances from large video collections. 
Rai \etal~\cite{Rai2025} investigate road-obstacle segmentation as a video segmentation task and study the use of vision foundation models to exploit temporal information.

For temporal association, we use causal mask-overlap matching without learned matching networks. This preserves the absence of task-specific training, although SAM\,2 and Depth Anything V2 remain pretrained learned models.
%
%
%
\section{Method Description}\label{sec:method}

\subsection{Problem Formulation and Motivation}
Let $\VEC{x}_t \in \I^{H \times W \times 3}$, $\I = [0,1]$, denote the RGB image at time $t \in \mathcal{T} = \{1, 2, \ldots, T\}$ where $T$ is the sequence length or $T=1$ if only single frames are available. Given a prompted segmentation model and a monocular depth estimator, the goal is to produce a dense per-pixel OOD score map $\VEC{s} \in \I^{H \times W}$ that assigns high values to obstacle pixels on the drivable surface and low values to road and ignored regions (background). 
An overview of our OOD segmentation and tracking approach is given in Fig.~\ref{fig:overview}.
 
Our key finding is that foundation segmentation networks already provide useful indications for OOD detection. 
We use this model to reliably segment the drivable road area and identify uncertain (i.e., low prediction scores) areas within the road as OOD. 
In this setting, the model does not require OOD-specific information or training examples, instead, it leverages the strong generalization capabilities it has acquired through large-scale pretraining. 
The segmentation process is controlled by various prompt types, including positive and negative point prompts as well as bounding boxes, enabling flexible adaptation to different scene configurations. 
However, the resulting predictions still tend to generate false positives in visually complex areas such as textured road surfaces or semantic scene boundaries.

To address these false positives, we incorporate (i) classical image processing tools to suppress small noise fragments as well as inaccurate edge predictions, and (ii) monocular depth estimation as an additional indicator of geometric consistency. 
The underlying assumption is that true OOD obstacles typically exhibit a measurable depth difference from the surrounding road surface, whereas many false positives (such as textured areas, road markings, or manhole covers) remain approximately colinear with the road. 
Based on this observation, we perform a segment-level verification of the associated OOD components by estimating their depth and comparing it to the depth distribution of the surrounding area. 
Candidate regions whose depth differs sufficiently from the local road context are retained as valid OOD objects, while geometrically consistent regions are suppressed as false positives. 
In this way, the monocular depth information complements the uncertainty-based predictions with an explicit geometric understanding of the scene.

\subsection{Pixel-level OOD Scores via Prompted Segmentation Model}
\label{sec:sam2}

\paragraph{Road segmentation.} 
We leverage a promptable foundation segmentation model as a zero-shot road segmenter. 
Various prompt strategies were investigated, including single-point prompts, multi-point prompts, and configurations with or without additional bounding box constraints. 
The following strategy has proven effective: 
The model receives a deterministic multi-point prompt: \emph{three} foreground points distributed horizontally across the lower portion of the frame, plus \emph{two} background points in the upper portion, constrained by a bounding box covering the bottom $70\%$ of the image. 
Positive prompts are placed in the lower central part of the image, corresponding to the road area directly in front of the vehicle, where the semantic association with the drivable surface is considered most reliable. 
In contrast, negative prompts are positioned in the upper part of the image, where buildings, vegetation, or areas of the sky are visible that should be excluded from the street prediction. 
Additionally, a generously sized bounding box is provided that covers the lower $70\%$ of the image. 
Since the exact extent of the road is not known in advance, the bounding box is deliberately made large to ensure that the entire drivable road surface is contained. \sv{The prompt coordinates and box ratio assume a typical forward-facing camera. A close obstacle can contaminate a positive prompt, and different camera geometries require adapted prompt positions.} 
The strategy is shown on the left side of Fig.~\ref{fig:overview}. 

\paragraph{Candidate selection.} 
Given an input image and the corresponding prompt, the segmentation model produces $K$ candidate masks $\{\VEC{m}_k\}_{k \in \mathcal{K}}$, $\mathcal{K} = \{1, \ldots, K\}$, together with predicted IoU-based mask quality scores $\{q_k\}_{k \in \mathcal{K}}$ and full-resolution logit maps $\{\VECG{\ell}_k\}_{k \in \mathcal{K}}$, where $\VEC{m}_k \in \{0,1\}^{H \times W}$ and $\VECG{\ell}_k \in \R^{H \times W}$. We select the best candidate $k^*$ satisfying geometric plausibility
\begin{align}
    k^* = \arg\max_{k \in \mathcal{K}} \; q_k \quad \text{s.t.} \quad 
    & \sum_{i \in \mathcal{I}} m_{k,i} > 0.1 \, H\cdot W \notag \\
    & \wedge\; \VEC{m}_k \text{ touches bottom} \enspace,
    \label{eq:candidate_selection}
\end{align}
where $i \in \mathcal{I} = \{1, \ldots, H \cdot W\}$ denotes the pixel indices. 
Among all candidate masks, we select the mask with the highest confidence that additionally satisfies two geometric constraints. 
First, the mask must cover at least $10\%$ of the image area in order to suppress small predictions. 
Second, the mask is required to touch the bottom image boundary, reflecting the assumption that the drivable road surface is connected to the ego-vehicle and therefore visible at the lower part of the image.
The proposed prompting strategy and geometric filtering jointly improve robustness to varying road geometries and perspective effects, while ensuring spatial consistency with the ego-vehicle position.

\paragraph{Dense OOD scoring from road logits.} 
Although the segmentation output is given as a binary mask, the associated logit map contains additional information about the model confidence. 
Thus,
we extract the full-resolution logit map $\VECG{\ell}_{k^*}$ and transform them into a dense OOD anomaly score via sigmoid inversion
\begin{align}
	s^{\mathrm{SAM}}_i = 1 - \sigma(\ell_{k^*,i}) \enspace , \quad i \in \mathcal{I} \enspace,
	\label{eq:sam_score}
\end{align}
where $\sigma(\cdot)$ is the sigmoid function. 
Pixels confidently classified as road have large positive logits and thus $s^{\mathrm{SAM}}_i \approx 0$, while pixels that the model considers anomalous/not road yield $s^{\mathrm{SAM}}_i \approx 1$. This provides a continuous score $\VEC{s}^{\mathrm{SAM}} \in \I^{H \times W}$ rather than a binary decision,
which captures the confidence structure of the predicted segmentation and provides richer information for subsequent OOD analysis.

\subsection{False Positive Suppression}
\label{sec:depth}

\begin{figure}[t!]

    \centering
    \includegraphics[width=\linewidth]{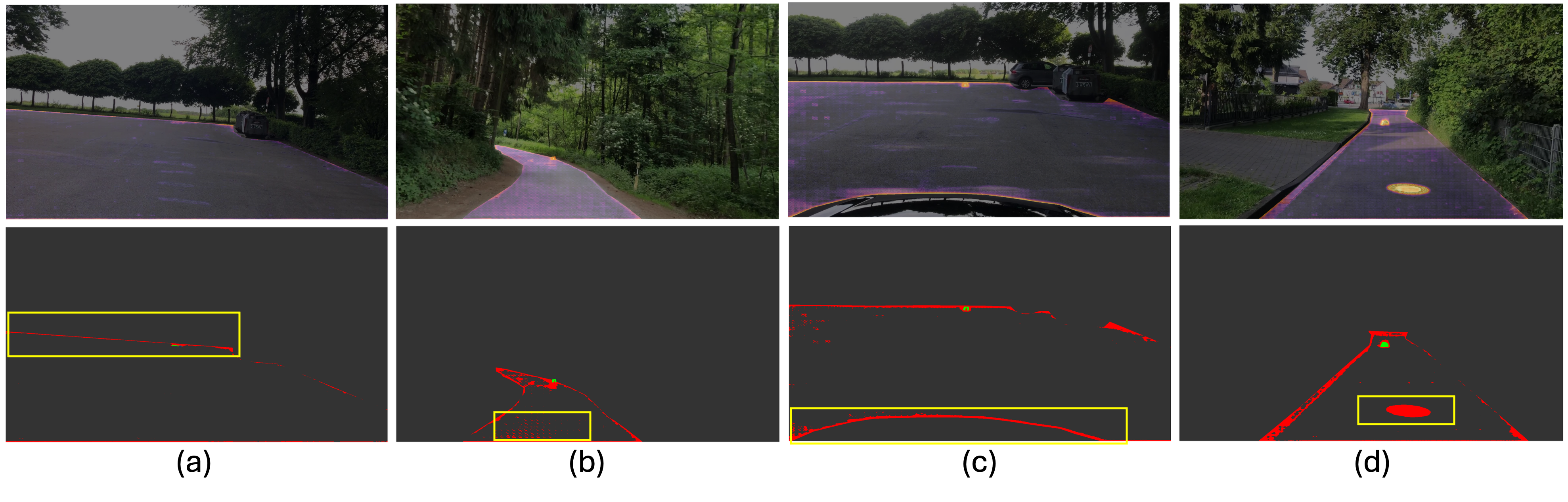}
    \caption{Examples of systematic false positives. Each column highlights a characteristic failure mode of the raw OOD predictions: (a) inaccurate road boundaries, (b) small isolated artifacts, (c) large connected components of ego-vehicle hood, and (d) flat road structures such as manhole covers. The first row shows the RGB image overlaid with the dense OOD score map. The second row visualizes the resulting OOD detections, where true positive OOD pixels are shown in green and false positive detections in red. }
    \label{fig:fp_strategie}
\end{figure}

The prediction uncertainties provided by the segmentation model already offer a strong signal for OOD detection. However, these predictions can be inaccurate at object edges, and certain structured road elements, such as manhole covers or lane markings, may be incorrectly identified as anomalous even though they are part of the road surface. In the following, we use depth estimation and classical image processing tools to reduce false positive OOD detections. 
In Fig.~\ref{fig:fp_strategie}, we show some examples and how our method suppresses these false positives.

\paragraph{From pixels to segments.} 
Instead of applying a pixel-wise false positive reduction, which we observed can break connected components and negatively affect segment-level evaluation, we perform filtering at the level of connected components (segments).

Given the dense OOD score $\VEC{s}^{\mathrm{SAM}} \in \I^{H \times W}$, we apply a fixed threshold $\theta = 0.5$ to obtain a binary mask $\mathcal{M}_{\mathrm{OOD}} \in \{0,1\}^{H \times W}$, which is subsequently processed using connected component analysis (CCA) to yield $N$ components $\{\mathcal{C}_n\}_{n \in \mathcal{N}}$ with $\mathcal{N} = \{1, \ldots, N\}$. 
We choose a constant threshold since the score distributions are typically highly bimodal due to the strong confidence separation of the underlying models, resulting in values that are either close to zero or close to one and thus showing limited sensitivity to the exact threshold choice.

\paragraph{Road mask expansion.} 
In practice, the prompted segmentation model tends to produce road masks that are slightly conservative at boundaries, undershooting curbs, sidewalk edges, and guardrails by a few pixels. 
As a consequence, OOD score values at these transitions can produce thin false-positive components along road edges that do not correspond to actual obstacles. 
To address this issue, we morphologically dilate the hole-filled road mask by $r_{\mathrm{road}} \in \N$ pixels
\begin{align}
	\mathcal{M}'_{\mathrm{road}} = \operatorname{dilate}(\mathcal{M}_{\mathrm{road}},\, r_{\mathrm{road}}) \enspace.
	\label{eq:road_dilate}
\end{align}
This expansion serves two purposes. 
First, candidate OOD segments that lie near the road boundary are now surrounded by valid road pixels, which enables their depth disparity to be properly evaluated in subsequent steps. 
Second, thin edge artifacts are absorbed into the expanded road region and therefore no longer appear as separate candidate components after thresholding. 
The dilation radius $r_{\mathrm{road}}$ is a hyperparameter that controls the trade-off between suppressing boundary-related false positives and preserving small obstacles located near road boundaries. In all experiments, we set $r_{\mathrm{road}}=10$ as the predicted road masks are already closely aligned with the actual road boundaries and only require minor expansion.

\paragraph{Heuristic segment filtering.} 
After connected component extraction, two classes of systematic false positives can be identified without requiring depth information. 
The first class consists of small noise fragments that arise from textured road surfaces or boundary artifacts. 
These components are typically very small in area and do not correspond to coherent objects. 
We therefore suppress all components with an area below a minimum threshold ($|\mathcal{C}_n| < A_{\min}$ pixels). 
We set $A_{\min}=200$ pixels, which corresponds to a very small fraction of the high-resolution input images and effectively filters out isolated noise while preserving meaningful object regions. 
The second class consists of large regions at the bottom of the image that correspond to the ego-vehicle hood, which consistently produces high OOD scores due to its visual dissimilarity from the road surface. 
Since legitimate road obstacles rarely span the full image width while simultaneously touching the vehicle's position, we suppress components that touch the bottom image border and exceed an area threshold ($|\mathcal{C}_n| \geq A_{\mathrm{bot}}$ pixels).
In practice, this constraint only affects a narrow line at the lower margin of the image and is therefore not sensitive to the exact choice of $A_{\mathrm{bot}}$ (we set $A_{\mathrm{bot}} = 500$). 
We emphasize that neither $A_{\min}$ nor $A_{\mathrm{bot}}$ were tuned on the evaluation datasets. Both values were fixed based on general considerations about typical noise fragment sizes and ego-vehicle regions, and the same values are used across all five evaluation datasets without modification.

\paragraph{Monocular depth estimation.} 
After heuristic filtering, the remaining false positives are components that are sufficiently large and positioned away from the image border, yet do not correspond to actual obstacles. 
These typically arise from flat road structures such as manhole covers, painted markings, or shadow transitions that produce strong appearance-based OOD scores. 
The key observation is that such false positives are geometrically flush with the surrounding road surface, whereas true obstacles protrude above or below the road plane. 
We acknowledge that this depth-based criterion is primarily effective for three-dimensional obstacles and may not suppress flat anomalous objects (e.g., debris or spills) that are coplanar with the road. However, such flat objects are typically less safety-critical for vehicle navigation than protruding obstacles, and they are already partially captured by the upstream appearance-based OOD scores.
Based on this observation, we incorporate monocular depth estimation as a geometric verification step.

We estimate per-pixel depth using a monocular depth foundation model, producing a dense depth map $\VEC{d} = (d_i) \in \R^{H \times W}_+$ for each frame. 
The depth map is normalized to the interval $[0,1]$ via per-frame min-max scaling, resulting in $\hat{\VEC{d}} \in \I^{H \times W}$, where larger values correspond to closer objects. 
No camera intrinsics or stereo calibration are required, since the verification relies only on relative depth differences within each frame.

\paragraph{Depth-ring scoring.} 
To evaluate whether a candidate segment is geometrically consistent with or distinct from the road surface, a local depth reference is created. 
A global road depth would be insufficient, since the road plane itself exhibits substantial depth variation due to perspective projection. 
We therefore construct a \emph{depth ring} $\mathcal{R}_n \subset \mathcal{I}$ for each remaining component $\mathcal{C}_n$ that captures only the immediate road context:
\begin{align}
	\mathcal{R}_n = \big(\operatorname{dilate}(\mathcal{C}_n, r) \;\cap\; \mathcal{M}'_{\mathrm{road}}\big) \;\setminus\; \mathcal{C}_n \enspace,
	\label{eq:ring}
\end{align}
where $r \in \N$ is the dilation radius and $\mathcal{M}'_{\mathrm{road}}$ is the expanded road mask from Eq.~\eqref{eq:road_dilate}. 
In this way, the ring contains only road pixels in the direct neighborhood of the candidate, providing a depth reference that naturally adapts to the local perspective and road geometry at the position of the component.

Each component is scored by the absolute depth disparity between its interior and the surrounding/neighboring ring
\begin{align}
	\Delta_n = \big|\operatornamewithlimits{median}_{i \in \mathcal{C}_n} \hat{d}_i - \operatornamewithlimits{median}_{i \in \mathcal{R}_n} \hat{d}_i\big| \enspace, \quad n \in \mathcal{N} \enspace.
	\label{eq:disparity}
\end{align}
The use of the median rather than the mean provides robustness against outlier pixels within both the component and the ring. 
All components whose disparity falls below a very small minimum threshold $\delta_{\min} \in \R^+_0$ are discarded, reflecting the assumption that they are geometrically consistent with the road plane and therefore unlikely to correspond to OOD objects. 
In practice, we set $\delta_{\min}=0.001$, such that only components with virtually no measurable depth deviation are removed, i.e., a true obstacle must exhibit a minimum depth discontinuity relative to its surroundings.


\paragraph{Temporal tracking.} 
To associate detected OOD instances across consecutive frames, we employ an online tracking strategy based on optimal bipartite matching combined with lifecycle state management~\cite{Bewley2016}. 
Given the set of retained components $\{\mathcal{C}_n\}$ in frame~$t$ and the tracked instances $\{\mathcal{T}_m\}$ from frame~$t{-}1$, we compute the intersection-over-union (IoU) between all pairs of current components and previous instances. We formulate the temporal association as a bipartite assignment problem and solve it using the Hungarian algorithm, minimizing total non-overlap with cost function $1 - \mathrm{IoU}$, which ensures globally optimal frame-to-frame correspondences. Each component is matched to a previous instance only if the IoU exceeds a minimum threshold $\tau_{\mathrm{IoU}}$. Components that cannot be matched to any existing instance enter a tentative state, and are confirmed as new tracks only after appearing in consecutive frames (controlled by parameter $n_{\mathrm{init}}$). This state-based filtering effectively suppresses transient false positive detections while maintaining stable object identities over time. Tracks are considered lost after remaining unmatched for a maximum number of frames ($\max\!_{\mathrm{age}}$), striking a balance between persistence and detection cleanup.

While SAM\,2 natively supports video-level mask propagation through its memory mechanism, we deliberately employ a separate online tracking strategy. This design choice ensures that the tracking component remains causal (using only past and current frames), which is a requirement for real-time deployment. Furthermore, decoupling detection and tracking allows each component to be independently evaluated and replaced, providing greater modularity. 
\sv{Overall, the proposed framework combines the generalization capabilities of pretrained foundation models with uncertainty-based OOD reasoning and geometric scene understanding. The resulting segmentation method is training-free and designed for road-centered obstacles, and the tracking component is a preliminary online association baseline.}

%
%
%
\section{Numerical Experiments}\label{sec:exp}

%
%
\subsection{Experimental Setting}\label{sec:exp_setting}

%
%

\paragraph{Models.} 
Our method relies on two foundation models that serve complementary roles in the proposed approach.
For road segmentation and the derivation of dense OOD scores, we employ SAM\,2~\cite{Ravi2025} in its large configuration.
SAM\,2 is a promptable segmentation model trained on a large-scale and diverse dataset, which enables it to generalize to unseen domains without task-specific adaptation.
Its ability to accept flexible input prompts, including point prompts and bounding boxes, makes it particularly suitable for our zero-shot road segmentation setting, where no training on driving-specific or OOD data is performed. 
For monocular depth estimation, we use Depth Anything V2~\cite{Yang2024} with a ViT-L backbone.
This model provides dense per-pixel depth predictions from a single RGB image. 
Both models are used exclusively with their publicly available pretrained weights, no fine-tuning or domain adaptation is applied, which is consistent with the training-free design of our overall approach.

\begin{table*}[t!]
\caption{Left: Ablation study on the prompting strategy for SAM\,2 on SOS and LostAndFound datasets, evaluated using pixel-level metrics AuPRC and FPR$_{95}$. 
Right: Ablation study on the cumulative effect of each post-processing stage on SOS and LostAndFound datasets, evaluated using segment-level $\overline{F_1}$-score.
}
\centering
\scalebox{0.85}{
\begin{tabular}{cl cc cc}
\toprule
& & \multicolumn{2}{c}{SOS} & \multicolumn{2}{c}{LostAndFound} \\
\cmidrule(lr){3-4} \cmidrule(lr){5-6}
Row & Configuration & AuPRC $\uparrow$ & FPR$_{95}$ $\downarrow$ & AuPRC $\uparrow$ & FPR$_{95}$ $\downarrow$ \\
\midrule
1 & SAM\,2, 1\,point & 7.05 & 100.00 & 4.48 & 11.30 \\
2 & SAM\,2, 3P+2N & 18.89 & 4.47 & 24.46 & \textbf{4.88} \\
3 & SAM\,2, 3P+2N+BB (70\%) & \textbf{93.34} & \textbf{0.22} & \textbf{71.13} & 3.20 \\
\bottomrule
\end{tabular}}
\hspace{1ex}
\scalebox{0.85}{
\begin{tabular}{cl c c}
\toprule
Row & Configuration & \multicolumn{1}{c}{SOS} & \multicolumn{1}{c}{LostAndFound} \\
\cmidrule(lr){1-2} \cmidrule(lr){3-3} \cmidrule(lr){4-4} 
4 & \quad + Road mask dilation & 2.93 & 48.87 \\
5 & \quad + Ego-vehicle suppression & 2.91 & 49.14 \\
6 & \quad + Small component filtering & 23.02 & 48.89 \\
7 & \quad + Depth-based suppression & \textbf{33.65} & \textbf{49.58} \\
\bottomrule
\end{tabular}}
\label{tab:ablation1}
\end{table*}
%
%
%
\begin{table}[t!]
\caption{OOD segmentation benchmark results on the test split of the LostAndFound dataset.}
\centering
\scalebox{0.85}{
\begin{tabular}{l cc ccc}
\toprule 
 & \multicolumn{5}{c}{LostAndFound} \\
\cmidrule(r){2-6} 
& AuPRC $\uparrow$ & FPR$_{95}$ $\downarrow$ & $\overline{\text{sIoU}}$ $\uparrow$ & $\overline{\text{PPV}}$ $\uparrow$ & $\overline{F_1}$ $\uparrow$ \\
\midrule
Maximum Softmax    & 30.1 & 33.2 & 14.2 & \textbf{62.2} & 10.3 \\
MC Dropout         & 36.8 & 35.6 & 17.4 & 34.7 & 13.0 \\
Ensemble           & 2.9 & 82.0 & 6.7 & 7.6 & 2.7 \\
PGN                & \underline{69.3} & \underline{9.8} & \textbf{50.0} & 44.8 & \underline{45.4} \\ 
W-EDL                & 50.4 & 25.8 & 21.2 & 38.0 & 19.2 \\
\midrule
ODIN               & 52.9 & 30.0 & 39.8 & 49.3 & 34.5 \\
Mahalanobis        & 55.0 & 12.9 & 33.8 & 31.7 & 22.1 \\
\midrule
Ours               & \textbf{71.5} & \textbf{3.2} & \underline{44.7} & \underline{52.6} & \textbf{49.6}\\
\bottomrule
\end{tabular} }
\label{tab:ood_laf}
\end{table}
%

%
%
\paragraph{Datasets.} 
To assess the OOD segmentation performance, we employ the three datasets provided by the SegmentMeIfYouCan benchmark\footnote{\url{http://segmentmeifyoucan.com/}}. 
The LostAndFound dataset \cite{Pinggera2016} contains 1,203 validation images with annotations for both, the road surface and OOD objects, mainly representing small obstacles located on German roads in front of the ego-vehicle. 
An updated variant, referred to as LostAndFound test-NoKnown, was later introduced in \cite{Chan2021_1} to further refine the evaluation protocol. 
The RoadObstacle dataset \cite{Chan2021_1} comprises 412 test images and follows a similar setup, where obstacles are also positioned on the road. Compared to LostAndFound, it provides a broader variety of obstacle types and environmental situations. 
The RoadAnomaly dataset \cite{Chan2021_1} contains 100 test images featuring diverse anomalous objects that may occur at arbitrary positions within the scene rather than being restricted to the road area. 

Moreover, we extend the evaluation to the Street Obstacle Sequences (SOS) and Wuppertal Obstacle Sequences (WOS) datasets\footnote{\url{https://rrow2024.github.io/challenge}}, which provide annotated video sequences of OOD objects over multiple frames \cite{Maag2022_OOD_tracking}. 
The SOS dataset contains 20 real-world video sequences recorded at a rate of 25 frames per second. The dataset includes 13 different OOD object categories and provides pixel-accurate annotations for every eighth frame, resulting in 1,129 labeled images.
The WOS dataset focuses on moving OOD objects and therefore introduces additional temporal challenges for tracking. It consists of 44 video sequences containing dynamic anomalies such as dogs, rolling balls, or skateboards, recorded with either static or moving cameras. Similar to SOS, every eighth frame is pixel-wise annotated, yielding a total of 938 labeled frames. 
%
%

\paragraph{Evaluation Metrics.} 
To evaluate OOD segmentation performance, we follow the SegmentMeIfYouCan benchmark protocol \cite{Chan2021_1}. 
On pixel-level, we report the area under precision-recall curve (AuPRC) and the false positive rate at 95\% true positive rate (FPR$_{95}$), measuring class separability and safety-critical false positive behavior, respectively. 
On segment-level, we use an adjusted version of the mIoU (sIoU), the positive predictive value (PPV), and the $F_1$-score. These metrics are averaged over thresholds ranging from 0.25 to 0.75 in steps of 0.05, resulting in $\overline{\text{sIoU}}$, $\overline{\text{PPV}}$ and $\overline{F_1}$. 
For OOD tracking, we additionally report the standard multi-object tracking metrics \cite{Bernardin2018}. MOTA (multiple object tracking accuracy) summarizes tracking performance by jointly accounting for false positives, false negatives, and identity mismatches, whereas MOTP (multiple object tracking precision) measures the spatial localization accuracy of matched object tracks. \sv{In our implementation, MOTP is the mean centroid distance in pixels over matched prediction-ground truth pairs. Consequently, lower values indicate more accurate localization.}

\paragraph{Baselines.} 
We primarily compare against established uncertainty-based OOD detection methods, i.e., Maximum Softmax, MC Dropout, Ensemble, PGN and W-EDL. 
In addition, we include adversarially enhanced approaches such as ODIN and Mahalanobis using input perturbations. 
These methods provide the most relevant baselines, as they operate directly on the predictions of segmentation networks without requiring external OOD datasets, extensive retraining, or auxiliary models. 
For OOD tracking, we additionally compare against existing state-of-the-art tracking approaches. However, it should be noted that the competing methods typically rely on considerably more complex pipelines, making a direct comparison difficult. 
%
%
\subsection{Numerical Results: Method Analysis}\label{sec:ablation}

\begin{table*}[t]
\caption{OOD segmentation benchmark results for RoadAnomaly and RoadObstacle datasets.}
\centering
\scalebox{0.85}{
\begin{tabular}{l cc ccc cc ccc}
\toprule 
\multicolumn{1}{c}{} & \multicolumn{5}{c}{RoadAnomaly} & \multicolumn{5}{c}{RoadObstacle} \\ 
\cmidrule(r){2-6} \cmidrule(r){7-11}
& AuPRC $\uparrow$ & FPR$_{95}$ $\downarrow$ & $\overline{\text{sIoU}}$ $\uparrow$ & $\overline{\text{PPV}}$ $\uparrow$ & $\overline{F_1}$ $\uparrow$ & AuPRC $\uparrow$ & FPR$_{95}$ $\downarrow$ & $\overline{\text{sIoU}}$ $\uparrow$ & $\overline{\text{PPV}}$ $\uparrow$ & $\overline{F_1}$ $\uparrow$\\
\midrule 
Maximum Softmax    & 28.0 & 72.1 & 15.5 & 15.3 & 5.4    
& 15.7 & 16.6 & 19.7 & 15.9 & 6.3 \\
MC Dropout         & 28.9 & 69.5 & 20.5 & 17.3 & 4.3    
& 4.9 & 50.3 & 5.5 & 5.8 & 1.1 \\ 
Ensemble           & 17.7 & 91.1 & 16.4 & \underline{20.8} & 3.4  
& 1.1 & 77.2 & 8.6 & 4.7 & 1.3 \\
PGN             & \underline{36.7} & \underline{61.4} & \underline{21.6} & 17.5 & \underline{6.2}  
& 16.5 & 19.7 & 19.5 & 14.9 & 7.4 \\
W-EDL             & \textbf{54.3} & \textbf{50.4} & \textbf{26.3} & \textbf{21.5} & \textbf{7.8} & \underline{30.7} & 42.3 & 11.6 & \underline{26.4} & 8.7 \\
\midrule
ODIN               & 33.1 & 71.7 & 19.5 & 17.9 & 5.2  
& 22.1 & 15.3 & \underline{21.6} & 18.5 & \underline{9.4} \\
Mahalanobis        & 20.0 & 87.0 & 14.8 & 10.2 & 2.7  
& 20.9 & \underline{13.1} & 13.5 & 21.8 & 4.7 \\
\midrule
Ours          & 14.6 & 69.7 & 10.3 & 9.0 & 5.4 
& \textbf{51.3} & \textbf{3.8} & \textbf{45.6} & \textbf{45.7} & \textbf{43.2}  \\
\bottomrule
\end{tabular} }
\label{tab:ood_ao}
\end{table*}
%
%
\begin{table*}[t]
\caption{OOD segmentation and tracking benchmark results for SOS and WOS datasets.}
\centering
\scalebox{0.85}{
\begin{tabular}{l cc ccc cc ccc}
\toprule 
\multicolumn{1}{c}{} & \multicolumn{5}{c}{SOS} & \multicolumn{5}{c}{WOS} \\ 
\cmidrule(r){2-6} \cmidrule(r){7-11}
& AuPRC $\uparrow$ & FPR$_{95}$ $\downarrow$ & $\overline{F_1}$ $\uparrow$ & MOTA $\uparrow$ & MOTP $\downarrow$ & AuPRC $\uparrow$ & FPR$_{95}$ $\downarrow$ & $\overline{F_1}$ $\uparrow$ & MOTA $\uparrow$ & MOTP $\downarrow$ \\
\midrule 
Baseline & 85.20 & 1.30 & 50.40 & 0.32 & 12.45 & \underline{94.92}  & 0.59             & 30.13             & 0.13             & 51.17 \\
PixOOD & \textbf{94.76} & \underline{0.18} & 47.15 & \underline{0.41} & 8.24       & \textbf{97.57}     & \textbf{0.21}    & \underline{43.36} & 0.00             & \underline{14.05} \\
JSR-Net & 87.01 & 1.59 & 9.98 & -4.59 & 97.67 & 59.66              & 19.49            & 2.11              & -19.30           & 301.52 \\
DaCUP & \underline{93.57} & \underline{0.17} & 20.25 & -1.34 & 89.79 & 83.13              & 3.77             & 12.93             & -2.26            & 280.74 \\
UNO-SAM2 & 90.21 & 0.18 & 15.45 & \textbf{0.44} & \textbf{2.15} & 88.10              & \underline{0.22} & 26.57             & \textbf{0.37}    & \textbf{3.98} \\
RbA & 89.47 & 0.33 & \underline{53.58} & 0.36 & 5.93 & 93.76              & 0.81             & \textbf{48.52}    & \underline{0.23} & 16.88 \\
\midrule
Ours & \underline{93.34} & 0.22 & 33.65 & -0.032 & 85.77 & 74.27              & 2.39             & 15.13				   & -0.029               & 407.55  \\
\bottomrule
\end{tabular} }
\label{tab:ood_sw}
\end{table*}

In this section, we describe the step-by-step development of our method, presenting the design decisions and the empirical observations that motivated each subsequent refinement. 
All ablation and method analysis experiments are conducted on the LostAndFound dataset, since the RoadAnomaly and RoadObstacle test sets are not publicly available. 
For the second benchmark, we primarily use the SOS dataset, as it provides a smaller and less complex task compared to WOS, making it more suitable for controlled analysis of the proposed framework.

\paragraph{OOD scores via prompted SAM\,2.} 
Table~\ref{tab:ablation1} (left) summarizes the results of investigated prompting configurations on the SOS and LostAndFound datasets.
Our investigation begins with the question of how to elicit useful OOD information from a foundation segmentation model without any task-specific training. The simplest approach is to provide the model with a single foreground point at the center bottom of the image, encoding a minimal prior that the road is located above the ego-vehicle (Row~1). However, this configuration produces poor pixel-level results on both datasets (AuPRC values below $7.05\%$), indicating that a single point does not provide sufficient context for the model to reliably identify the road region.

We therefore enrich the prompt by distributing three positive points horizontally across the lower image region and adding two negative points in the upper portion (Row~2). This configuration explicitly communicates the spatial structure of street scenes, i.e., road below, non-road above, to the model. The resulting AuPRC improves substantially ($18.89\%$ on SOS, $24.46\%$ on LostAndFound), confirming that negative prompts help the model suppress ambiguous predictions in sky and building regions.

Despite this improvement, the pixel-level scores remain insufficient for reliable OOD detection. A qualitative inspection of the logit maps reveals that, without spatial constraints, the model occasionally assigns high confidence to non-road regions that happen to share textural similarity with the road. To address this, we additionally provide a bounding box covering the lower $70\%$ of the image (Row~3). This constraint restricts the model's attention to the plausible road area and produces substantially sharper logit maps. The resulting AuPRC reaches $93.34\%$ on SOS and $71.13\%$ on LostAndFound, establishing a strong pixel-level baseline from which to derive segment-level predictions.

\paragraph{False positive suppression.} 
Table~\ref{tab:ablation1} (right) summarizes the impact of the different false-positive suppression stages on the SOS and LostAndFound datasets. 
As the underlying pixel-wise OOD scores are identical for Rows~4--7, the pixel-level metrics remain unchanged, while the impact of the segment-level refinements becomes visible in the $\overline{F_1}$ score. 

As a first step, we threshold the dense scores and apply connected component analysis within the dilated road mask to obtain initial segment candidates (Row~4). On LostAndFound, where obstacles are typically well-separated from road boundaries, this already yields a $\overline{F_1}$ of $48.87\%$. On SOS, however, the result is poor ($2.93\%$), as the candidate set is dominated by false positives, in particular, large segments corresponding to the ego-vehicle hood and numerous small noise fragments along road edges.

An analysis of the false positive patterns on SOS reveals two recurring sources of error: (i)~the ego-vehicle hood, which consistently produces a large high-scoring region at the bottom of the image, and (ii)~small scattered fragments arising from textured road surfaces and boundary artifacts. 
To address the first issue, we suppress large connected components ($\geq A_{\mathrm{bot}}$ pixels) that touch the bottom image boundary (Row~5). This heuristic is motivated by the observation that legitimate road obstacles rarely span the full image width while simultaneously touching the vehicle's position. 
For the second issue, we remove components below a minimum area threshold $A_{\min}$ (Row~6). On SOS, this step produces the most substantial single improvement in $\overline{F_1}$, from $2.91\%$ to $23.02\%$, confirming that the majority of initial false positives are small noise fragments rather than coherent object detections. 

After heuristic filtering, the remaining false positives mainly correspond to flat road structures such as manhole covers, lane markings, or shadow transitions that produce high appearance-based OOD scores despite being geometrically consistent with the road surface. 
To suppress these cases, the depth-ring scoring mechanism (Row~7) compares the median depth of each candidate component to the surrounding road region. Components whose depth disparity $\Delta_n$ falls below the threshold $\delta_{\min}$ are suppressed as geometrically consistent with the road plane. 
On SOS, this improves the $\overline{F_1}$ from $23.02\%$ to $33.65\%$. On LostAndFound, the $\overline{F_1}$ increases from $48.89\%$ to $49.58\%$, indicating that depth verification provides a complementary signal even on datasets where heuristic filtering is already effective.

\paragraph{Summary of findings.} 
The incremental development reveals that our method is built upon two complementary ideas. 
First, an appropriate prompting strategy already enables strong pixel-level OOD predictions from a foundation segmentation model without any additional training, with the bounding box constraint providing particularly important spatial context. Second, reliable segment-level predictions require suppressing systematic false positives through geometric heuristics and depth-based verification. The final configuration (Row~7) combines these components into the complete proposed framework compared in the following section with baselines.

\subsection{Numerical Results: Benchmarking}\label{sec:exp_results}
First, we compare our method with the comparable baseline models from the SegmentMeIfYouCan benchmark, i.e., with uncertainty-based approaches (Maximum Softmax, MC Dropout, Ensemble, PGN and W-EDL) as well as with those based on adversarial attacks (ODIN and Mahalanobis). The results are given in Table~\ref{tab:ood_laf} for LostAndFound, and in Table~\ref{tab:ood_ao} for RoadAnomaly and RoadObstacle. 
While our method achieves weaker performance on RoadAnomaly, this result directly reflects the deliberate scope of our approach: the method is designed for road-surface obstacles and does not claim to detect OOD objects at arbitrary scene positions. RoadAnomaly includes anomalies on sidewalks, in vegetation, and in the sky, which fall outside the drivable area targeted by our road segmentation. In contrast, our proposed framework achieves strong results on LostAndFound and especially on RoadObstacle, where road-centered OOD objects are predominant. On RoadObstacle, our method reaches a pixel-level AuPRC of 51.3 compared to 30.7 for the second-best approach (W-EDL), and a segment-level $\overline{F_1}$ score of 43.2 compared to only 9.4 for the next strongest baseline (ODIN). Notably, the proposed training-free framework also outperforms substantially more complex methods such as JSRNet and Image Resynthesis, which rely on auxiliary reconstruction or discrepancy networks.

Second, we evaluate our method on the OOD tracking benchmarks SOS and WOS, with the corresponding results reported in Table~\ref{tab:ood_sw}. As already discussed in the related work section, considerably fewer methods have been evaluated on these benchmarks, and comparable uncertainty-based OOD approaches are currently not available. Instead, the existing baselines rely on substantially more complex frameworks, including the use of additional OOD training data (Baseline~\cite{Maag2022_OOD_tracking}), auxiliary models (JSRNet, DaCUP and  UNO-SAM2), latent feature-space representations (PixOOD), or retrieval-based strategies (RbA). 
For the OOD segmentation metrics, we achieve competitive or superior performance compared to more complex approaches such as the Baseline, DaCUP and RbA, while fully outperforming JSR-Net across all evaluated segmentation metrics. 
\sv{For the OOD tracking metrics, the simple IoU-based association yields limited tracking accuracy, particularly on WOS where dynamic OOD objects exhibit large inter-frame displacements that exceed the overlap-based matching capability. The negative MOTA values and large MOTP values in Table~\ref{tab:ood_sw} therefore identify temporal association as a limitation of the current pipeline, rather than evidence of competitive tracking performance.}


Across the SegmentMeIfYouCan benchmark, our method achieves strong performance on datasets where OOD objects appear on the road surface and is competitive with more complex methods. 
\sv{On the OOD tracking benchmarks (SOS and WOS), the pixel-level and segment-level detection metrics are competitive with several methods that rely on additional OOD training data, auxiliary models, or retrieval-based components. The tracking metrics, however, indicate room for improvement through more advanced temporal association strategies.} 
\sv{Overall, these results support the proposed training-free OOD segmentation framework for road-centered obstacle detection in driving scenarios. 
}


\section{Conclusion}\label{sec:conc}

\sv{We have presented a training-free approach for road-centered OOD segmentation, together with a preliminary online tracking extension, that combines the confidence predictions of a foundation segmentation model with monocular depth estimation. Our key finding is that modern pretrained foundation models, when prompted appropriately, can provide useful information for distinguishing road surfaces from unknown obstacles without OOD-specific training data or task-specific fine-tuning.}
By inverting the dense logit maps of SAM\,2 into continuous anomaly scores, strong pixel-level OOD detection is achieved through a carefully designed prompting strategy alone.

To transition from pixel-level scores to reliable segment-level predictions, we have introduced a depth-ring scoring mechanism that exploits the geometric structure of driving scenes.
The underlying principle, i.e., that true obstacles extend beyond the road plane while appearance-based false positives remain geometrically flush with it, provides a complementary verification signal that is independent of the visual appearance of the scene.
Combined with heuristic filtering of systematic false positive patterns, this enables competitive segment-level performance without learned post-processing components.

\sv{Experimental evaluation on five datasets from two established benchmarks demonstrates that the proposed segmentation method achieves results comparable to or exceeding those of methods that rely on dedicated training procedures, OOD data, or complex auxiliary models.} 
\sv{In particular, on RoadObstacle, our approach outperforms all compared uncertainty-based and adversarial baselines on both pixel-level and segment-level metrics, while on SOS it achieves the second-highest AuPRC among all methods. The tracking results should be interpreted as a first causal baseline and motivate future work on association under large motion.}


\bibliographystyle{IEEEtran}
\bibliography{egbib}

\end{document}